\documentclass[runningheads]{llncs}
\usepackage{graphicx}

\usepackage{float}
\usepackage{subfigure}
\usepackage{graphicx}
\usepackage{hyperref}

\usepackage[pscoord]{eso-pic} %necessary for the watermark / information about upcoming publication 

\usepackage[export]{adjustbox}
\begin{document}
\title{Explainable Machine Learning for Fraud Detection}
\author{Ismini Psychoula \and
Andreas Gutmann \and
Pradip Mainali \and S. H. Lee  \and Paul Dunphy  \and Fabien A. P. Petitcolas}
\authorrunning{I. Psychoula et al.}
\institute{OneSpan Innovation Centre, Cambridge, UK \\
\email{\{first.last\}@onespan.com}}
\maketitle              % typeset the header of the contribution

\AddToShipoutPictureBG*{
    \put(\LenToUnit{.5\paperwidth},\LenToUnit{.955\paperheight}) % parameters to define vertical and horizontal reference point of text, respectively
    {\vtop{{\null}\makebox[0pt][c]{ % change "c" to "l" or "r" if you wish to align the text in a corner; don't forget to change the parameters above to define the reference point in the respective corner
        \small\sc\textcolor{gray}{To be published in IEEE Computer Special Issue on Explainable AI and Machine Learning} %define text (almost) arbitrarily; make italic, bold, small caps, etc. whatever you wish
    }}}
}

\begin{abstract}
The application of machine learning to support the processing of large datasets holds promise in many industries, including financial services. However, practical issues for the full adoption of machine learning remain with the focus being on understanding and being able to explain the decisions and predictions made by complex models. In this paper, we explore explainability methods in the domain of real-time fraud detection by investigating the selection of appropriate background datasets and runtime trade offs on both supervised and unsupervised models.

\end{abstract}
\section{Introduction}

%\blfootnote{To be published in IEEE Computer Special Issue on Explainable AI and Machine Learning.}

The digital landscape is constantly evolving and shifting towards integration of AI and machine learning in core digital service functionalities. Due to the Covid-19 pandemic, this has led to a shift in digital services, transforming from being a convenience to being a necessity. Many organizations had to transition to online services faster than anticipated. While this creates opportunities for development and growth, it also attracts cybercriminals.
Increases in the hacking of personal accounts and online financial fraud were reported during the strictest times of the lockdown during the Covid-19 outbreak~\cite{buil2020cybercrime} as well as more successful cybercrime by established and new threat actors, that could introduce new crime patterns~\cite{vu2020turning}. Fraud is costing businesses and individuals in the UK £130 billion annually and £3.89 trillion in the global economy~\cite{fraudcost}.
There is an increasing need for financial fraud detection systems that can automatically analyze large volumes of events and transactions and machine learning based risk analytics is one of the ways to achieve that. 

Many approaches have been proposed for automatic fraud detection, from anomaly detection to classical machine learning and modern deep learning models; however, it remains a challenging problem. There is no perfect universal rule to distinguish a fraudulent case from a valid one, as fraud appears in all shapes and sizes and can be indistinguishable from normal cases.
%% Added part for revision 
The work presented in~\cite{varmedja2019credit} compared Logistic Regression, Naive Bayes, Random Forests, Multilayer Perceptron, and Artificial Neural Networks. They found that Random Forests had the best performance. The authors also state the importance of using sampling methods to address the class imbalance issues that are common in fraud datasets. On a similar study presented in~\cite{thennakoon2019real} the authors offer guidance on the selection of optimal algorithms based on four types of fraud. A recent study proposed Discrete Fourier Transform conversion to exploit frequency patterns instead of canonical ones~\cite{saia2017frequency}. Another recent work explored the use of Prudential Multiple Consensus, which combines the results of several classification models based on the classification probability and majority voting~\cite{carta2019fraud}. 
%%%%

Another challenging aspect is that when applying complex models to detect fraud cases there is no easy way to explain how these methods work and why the model makes a decision. Unlike linear models like Logistic Regression, where the coefficient weights are easy to explain, there is no simple way for assessing the reasons behind a complex machine learning model's or deep neural network’s prediction. In particular, for applications with sensitive data or applications in safety critical domains, providing effective explanations to the users of the system is paramount~\cite{adadi2018peeking} and has become an ethical and regulatory requirement in many application domains~\cite{gdpr}. 

Explainability is not only linked to understanding the inner workings and predictions of complex machine learning models but also to concerns over inherent biases or hidden discrimination, potential harms to privacy, democracy, and other societal values. The General Data Protection Regulation (GDPR) states in Articles 13-14 and 22 that data controllers should provide information on `the existence of automated decision-making, including profiling' and `meaningful information about the logic involved, as well as the significance and the envisaged consequences of such processing for the data subject'. There are several important elements to consider, and be able to explain, when creating automated decision making systems: (1) The rationale behind the decision should be easily accessible and understandable. (2) The system should maximize the accuracy and reliability of the decisions. (3) The underlying data and features that may lead to bias or unfair decisions. (4) The context in which the AI is deployed and the impact the automated decision might have for an individual or society~\cite{europa}.

Several studies have been proposed to explain models in anomaly detection settings. Contextual Outlier INterpretation ~\cite{liu2017contextual} is a framework designed to explain anomalies spotted by detectors. Situ is another system for detecting and visualizing anomalies in computer network traffic and logs~\cite{goodall2018situ}, while in~\cite{collaris2018instance} the authors developed dashboards that provide explanations for insurance fraud detected by a Random Forest algorithm. Similar studies to this one have also used SHapley Additive exPlanation (SHAP) values for Autoencoder explanations~\cite{antwarg2019explaining},  explained network anomalies with variational Autoencoders~\cite{nguyen2019gee} and compared SHAP values to the reconstruction error of Principal Component Analysis features to explain anomalies~\cite{takeishi2019shapley}. However, so far there has not been a lot of focus on the impact of the background dataset and the run-time implications which such explanations could have for real time systems, such as fraud.

In this work, we explore explanations with two of the most prominent methods, LIME and SHAP, to explain fraud detected by both supervised and unsupervised models. Attribution techniques explain a single instance prediction by ranking the most important features that affected the generation of it. LIME~\cite{ribeiro2016should}  approximates the predictions of the underlying black box model by training local surrogate models to explain individual predictions. Essentially, LIME modifies a single data sample by tweaking the feature values in the simpler local model and observes the resulting impact on the output. The SHAP~\cite{lundberg2017unified} method explains the prediction of an instance by computing the contribution of each feature to the prediction using Shapley values based on coalition game theory. Intuitively, SHAP quantifies the importance of each feature by considering the effect each possible feature combination has on the output. The explanations aim to provide insights to experts and end users by focusing on the connection and trade offs between the features that affect the final decision depending on the background dataset and the run-time of the explanation method. We focused on black-box explanation methods because they can be applied to most algorithms without being aware of the exact model.

\section{Case Study} 

We present a financial fraud explainability case study. We use the open source IEEE-CIS Fraud Detection dataset~\cite{ieeefraud} to provide fraud detection explanations. The dataset provides information on credit card transactions and customer identity - with labels for fraudulent transactions ($Y \in {0,1}$). The dataset has highly imbalanced classes, with fraud accounting for 3.49\% of all transactions.

We experimented with both supervised and unsupervised models and compared their performance on the same dataset in terms of prediction accuracy, reliability of explanation, and run-time. 
For the supervised models, we used the labels provided in the IEEE-CIS Fraud Detection dataset~\cite{ieeefraud} to indicate for each transaction whether it is fraudulent or genuine during the training phase. 
However, obtaining labels for each transaction is often not possible and labeling the data manually or having just clean data is often difficult and time consuming. Unsupervised methods and representation learning can handle well imbalanced datasets without requiring labels. 
In the unsupervised models (Autoencoder and Isolation Forest) we treated the IEEE-CIS Fraud Detection dataset as unlabeled and used the reconstruction loss and anomaly score respectively to detect the fraud cases.  

We compared the following models:
\begin{itemize}
\item Naive Bayes: Probabilistic classifier that is simple, highly scalable and easy to train in a supervised setting.
\item  Logistic Regression: Simple and inherently intelligible model. This shows how much we gain by using more complex models compared to simple ones.  
\item Decision Trees: Decision trees offer robust accuracy and inherent transparency when their size is small.
\item Gradient Boosted Trees: Tree ensembles are one of the most accurate types of models but also quite complex. We trained the model with 100 estimators, maximum depth of 12 and learning rate of 0.002.  
\item Random Forests: An classifier that uses ensembles of trees to reduce predictive error. We trained the model with 100 estimators.
\item Neural Network: A multi-layer perceptron that can model non-linear interactions between the input features.  We trained the multi-layer perceptron with 3 hidden layers containing 50 units each with ReLU activation using the Adam optimizer. 
\item Autoencoder: An unsupervised neural network that works by using backpropagation and setting the target values to be equal to the inputs. We trained the network with 3 hidden layers containing 50 units each  with ReLU activation, Adam optimizer and Mean Squared Error (MSE) as the loss. The reconstruction error measures whether an observation deviates from the rest. 
\item Isolation Forest: An unsupervised algorithm that uses a forest of decision trees to partition the data. The splits to separate the data are done randomly. The number of splits indicates whether a point is an anomaly. For the training, we used 100 estimators and automatic contamination.
\end{itemize}

We only use 24 out of the 433 features in the dataset. The 24 features were selected to focus on the columns that have some description of their values so that the explanation could be more understandable. These features are `TransactionAMT' transaction payment amount in USD, `ProductCD'  the product for each transaction, `Device Type' and `Device Information', `card1' to `card6' which show payment card information, such as card type, card category, issuing bank, country etc.  `P\_emaildomain' is the purchaser's email domain and `R\_emaildomain' is the recipient's email domain,  `M1' to `M9' indicate matches, such as names on card and address, `id\_x' are numerical features for identity, such as device rating, IP domain rating, proxy rating etc. The dataset includes behavioral fingerprints like account login times and failed login attempts, as well as how long an account stayed on the page. However, the providers of the dataset were not able to elaborate on the meaning of all the features and correspondence between features and columns due to security terms and conditions~\cite{ieeefraud}.

\begin{table}[tbh]
\centering
\setlength{\tabcolsep}{3pt}
\caption{Performance Results}
%\addtolength{\tabcolsep}{2pt}
\begin{tabular}{ccccc}
	\hline
Model & Precision & Recall & F1-score & AUC  \\
	\hline
Naive Bayes	&  $0.543$  &  $0.669$ &  $0.544$ & $0.663$ \\
Logistic Regression & $0.891$ & $0.533$ & $0.553$ & $0.533$ \\
Decision Tree & $0.762$ & $0.742$ & $0.752$ & $0.706$\\
Random Forest &   $0.840$ & $0.725$ & $0.769$ & $0.688$	 \\
Gradient Boosting & $0.880$ & $0.729$  & $0.789$ &  $0.709$ \\
Neural Network& $0.795$	& $0.578$ & $0.619$ & $0.581$\\
Autoencoder	& $0.944$ & $0.767$ &  $0.839$  & $0.617$ \\
Isolation Forest & $0.723$  & $0.608$ & $0.664$ & $0.553$ \\
	\hline
\end{tabular}
\label{tab:supervised_models_report}
\end{table}

Table~\ref{tab:supervised_models_report} shows the classification results for each of the models. We withheld 20\% of the sample for validation. For the implementation of the models the Scikit-learn \footnote[1]{\url{http://scikit-learn.sourceforge.net/}} and Keras \footnote[2]{\url{https://keras.io/}} libraries were used.  We evaluated the performance of the models using precision, recall, F1-score and area under receiver operating characteristic curve, since the dataset is highly imbalanced.

\subsection{Trustworthiness of Explanations}

\begin{figure*}
\centering
\includegraphics[trim= 0 0 0 30pt, clip, width=\textwidth]{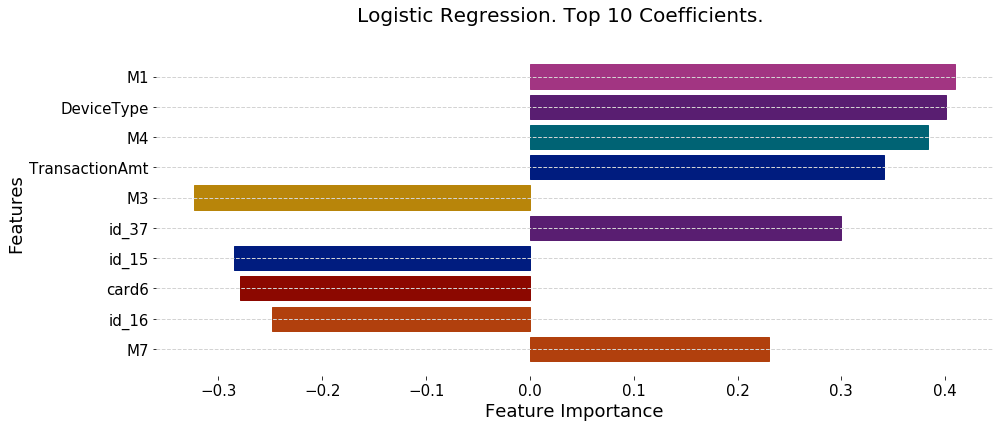}
\caption{Top 10 global features for Logistic Regression ranked by importance.}
\label{fig:log_reg}
\end{figure*}

To create a benchmark for the explanations we used a logistic regression classifier to predict fraudulent transactions and measure feature importance through the coefficient weights. Due to the transparency of the logistic regression model and its wide acceptance among regulatory bodies, we treat the global weights this provides as the ground truth and compare it with the results of attribution methods. Figure~\ref{fig:log_reg} presents the global top 10 most important features as determined by Logistic Regression.

\subsection{Explaining Supervised Models}

Attribution techniques such as LIME and SHAP explain a single instance prediction by ranking the most important features that affected the generation of it.
To evaluate the performance of LIME and SHAP in fraud detection, we compare and evaluate them by providing explanations with feature importance. We use summary plots that provide an overview of how the values of the same single instance have influenced the prediction for the different models. As single instance we define one fraudulent transaction that is used for all models and experiments.

\begin{figure*}[h]
\centering
\subfigure[LIME]{%
\label{fig:lime_imbalanced}%
\includegraphics[trim=8 5 10 35 ,clip, width=\textwidth]{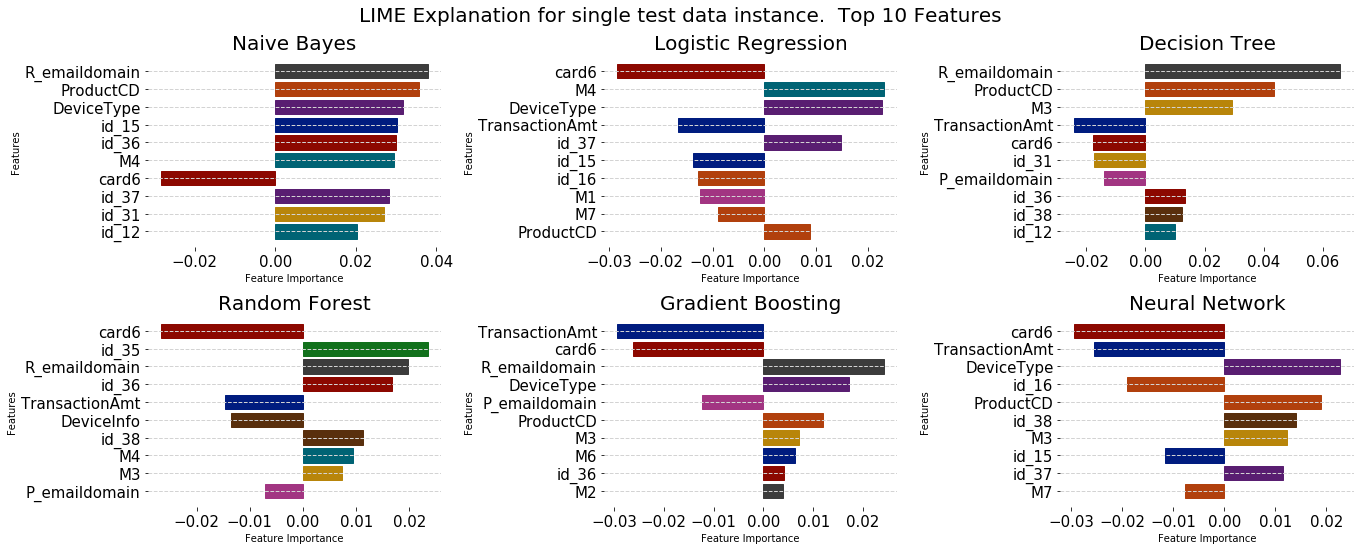}}\\
\vspace{8pt}%
\subfigure[SHAP]{%
\label{fig:shap_imbalanced}%
\includegraphics[trim=8 5 10 35, ,clip,width=\textwidth]{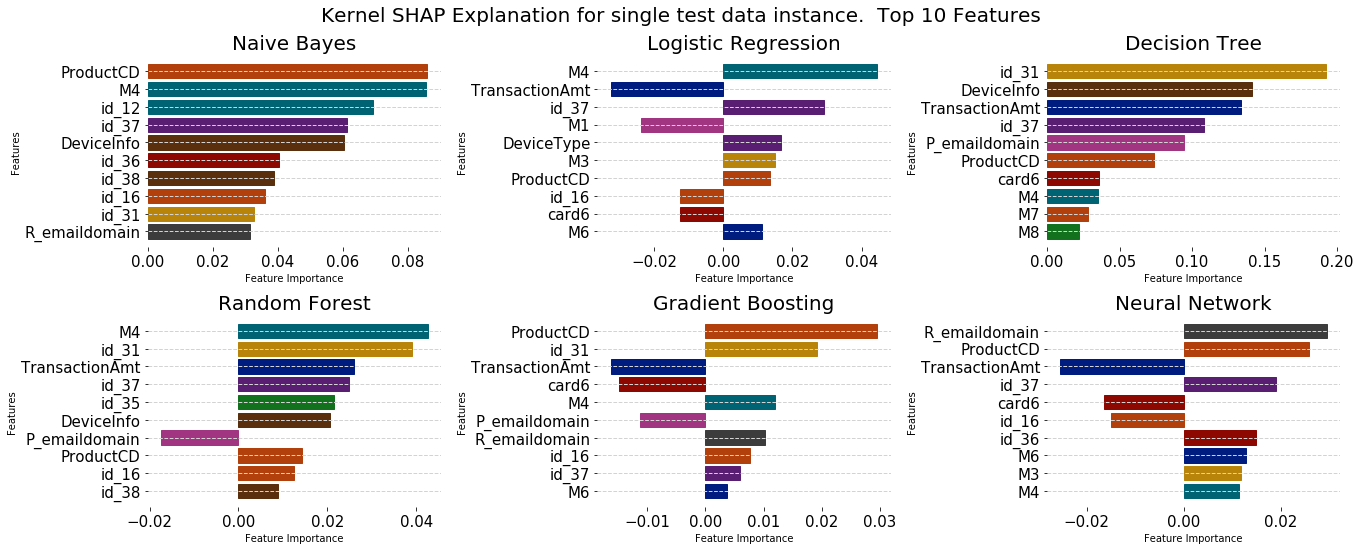}} 
\caption[Comparison between Explanations]{Top 10 features ranked by importance for the same normal instance with: 
\subref{fig:shap_normal} LIME
\subref{fig:shap_fraud} SHAP
}%
\label{fig:comparison_imbalanced}%
\end{figure*}

Experimental results for the supervised models are shown in Figure~\ref{fig:comparison_imbalanced}. Both LIME and SHAP produce similar top features with different rankings. Compared with the global features of Logistic Regression in Figure~\ref{fig:log_reg}, LIME agrees on 7 and SHAP agrees on 8 features. However, the ranking of features varies across both methods and all models. We noticed that on average SHAP produces explanations closer to the global features of Logistic Regression in terms of rank. 

\subsection{Explaining Unsupervised Models}

\begin{figure*}
\centering
\subfigure[Autoencoder]{%
\label{fig:autoencoder_shap}%
\includegraphics[trim=0 5 10 30, ,clip,width=0.49\textwidth]{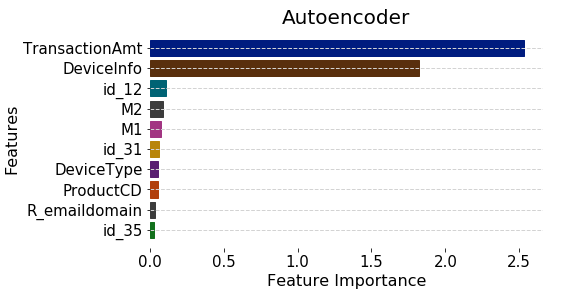}}
\vspace{8pt}%\\
%\hspace{15pt}
\subfigure[Isolation Forest]{%
\label{fig:isolation_shap}%
\includegraphics[trim=0 5 10 30, clip,width=0.49\textwidth]{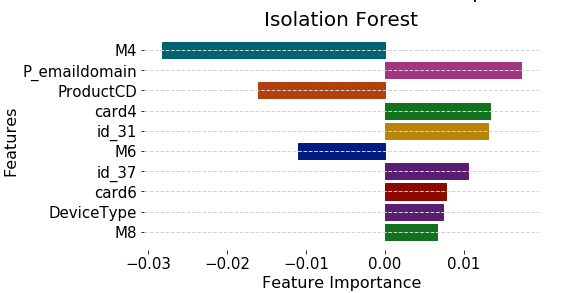}} 
\caption[Unsupervised SHAP Explanations]{Top 10 SHAP features ranked by importance for a single instance with: 
\subref{fig:shap_normal} Autoencoder
\subref{fig:shap_fraud} Isolation Forest
}%
\label{fig:comparison_unsupersvised_imbalanced}%
\end{figure*}

In unsupervised methods, particularly for anomaly detection, the result given by a model is not always a probability. In the case of the Autoencoder, we are interested in the set of explanatory features that explains the high reconstruction error. In Figure~\ref{fig:autoencoder_shap} we see that SHAP agrees only on 4 features, with Transaction Amount and Device Information the most important features that flagged the transaction as fraud.
Figure~\ref{fig:isolation_shap} shows the top explanation features for Isolation Forest with SHAP. We notice that the top features match only 4 of the explanations of Logistic Regression and 5 of the features from the Autoencoder.

\subsection{Reliability of Explanations}

The SHAP method requires a background dataset as a reference point to generate single instance explanations. In image processing, for example, it is common to use an all black image as a reference point but in financial fraud detection, there is no universal reference point that can be used as a baseline. We explored the impact of different background datasets in fraud detection explanations and evaluated different reference points that could be used to provide contrasting explanations to fraud analysts.

\begin{figure*}[h]
\centering
\subfigure[Normal Transactions]{%
\label{fig:shap_normal}%
\includegraphics[trim=0 0 0 35 ,clip,width=\textwidth]{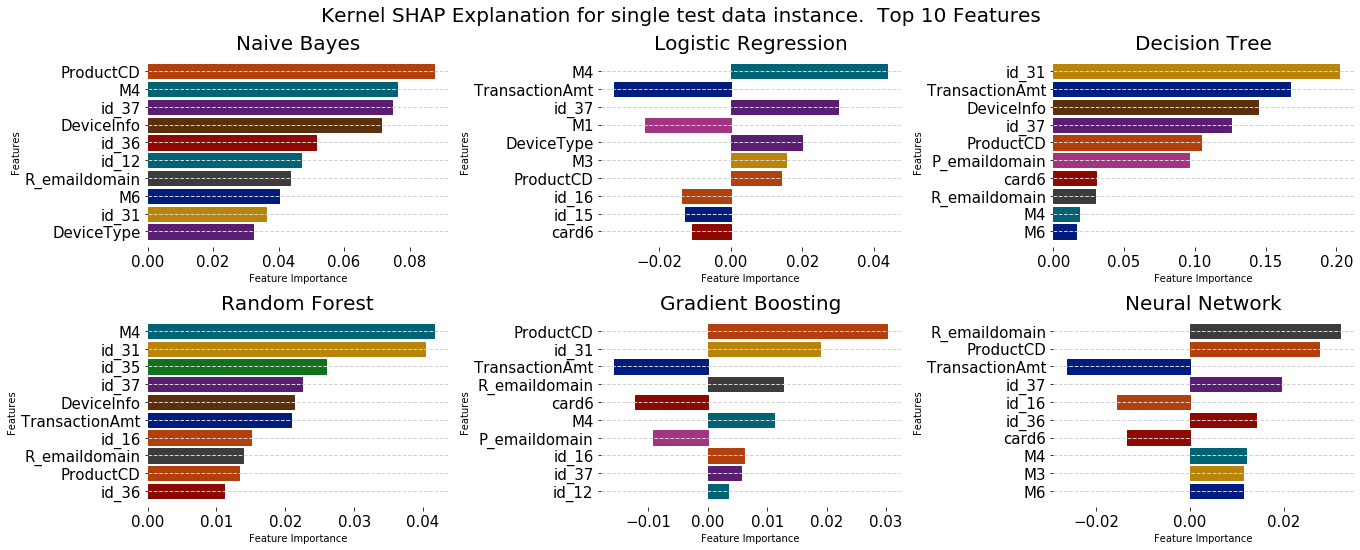}}\\
%\vspace{8pt}%
\subfigure[Fraud Transactions]{%
\label{fig:shap_fraud}%
\includegraphics[trim=0 0 0 35 ,clip,width=\textwidth]{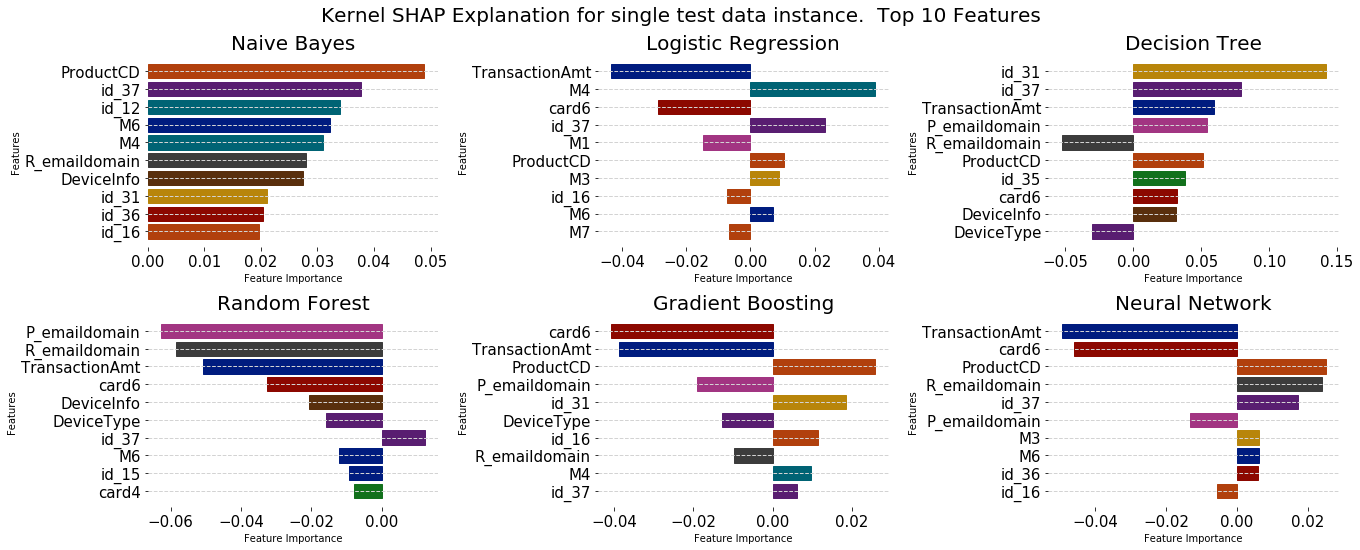}} 
\caption[Reliability of Explanations]{Top 10 features for single instance with SHAP using as background: 
\subref{fig:shap_normal} only Normal Transactions
\subref{fig:shap_fraud} only Fraud Transactions
}%
\label{fig:shap_background}%
\end{figure*}

Figure~\ref{fig:shap_background} highlights the differences when using only normal or only fraud transactions as reference point. We noticed that models like Naive Bayes, Logistic Regression and Decision Trees give more consistent explanations regardless of the background dataset, while models like Random Forest, Gradient Boosting and Neural Networks are more sensitive to the reference point. We also experimented with different background datasets for the Autoencoder and Isolation Forest models. %(due to space limitations we do not include the visualizations). 
Our findings show that the Autoencoder is more robust to changes in the background dataset. In the Isolation Forest model we find that contributing features remain mostly the same but their ranking is affected the most. 

We noticed that most models are sensitive to the choice of reference but there is no obvious point that can be used as a reference for fraud detection. By using an intuitive reference point we can provide a foundation upon which we can produce explanations that are either similar or contrasting (e.g. existing blacklisted accounts) to the class we are trying to predict. For each domain, it is important to understand which references are most understandable, trustworthy, and reliable for the end user.

Another important trade off to consider in real time systems is the time needed to provide an explanation. As we discussed previously, SHAP depends on background datasets to infer an expected value. For large datasets, it is computationally expensive to use the entire dataset and we rely on approximations (e.g. sub-sample of the data). However, this has implications for the accuracy of the explanation. Typically, the larger the background dataset the higher the reliability of the explanation. Table~\ref{tab:runtime} shows the time needed to provide a single instance explanation with SHAP based on different sub-sampled background sizes and with LIME. These experiments where run on a Linux server with an Asus TURBO RTX 2080TI 11GB GPU. For fraud detection systems that operate in real time, the fine tuning of the model and the explanation method will affect how many explanations a fraud analyst can receive in time. 

\begin{table*}[tbh]
\centering
\setlength{\tabcolsep}{6pt}
\caption{Run time for single instance explanation (in seconds), where $s$ is size of the sub-sampled background dataset.} 
%\addtolength{\tabcolsep}{2pt}
\begin{tabular}{ccccc}
\hline
Model &SHAP($s=600$) &SHAP($s=1000$) &SHAP($s=4000$) & LIME \\
\hline
Naive Bayes	& $4.32$ & $7.03$ & $30.61$  &$4.38$\\
Logistic Regression & $3.78$ & $6.43$ & $26.38$ & $4.43$ \\
Decision Tree & $3.88$ & $6.23$ & $27.55$ & $4.42$\\
Random Forest & $22.66$ & $35.67$ & $221.74$ &$4.55$\\
Gradient Boosting & $119.98$ & $193.31$ & $241.80$   & $5.19$ \\
Neural Network & $6.34$ & $11.10$ & $33.78$ &$4.44$\\
Autoencoder & $9.26$ & $14.66$ & $73.88$ & - \\
Isolation Forest & $39.11$ & $71.97$ & $318.59$ & - \\

	\hline
\end{tabular}
\label{tab:runtime}
\end{table*}

Overall, LIME and SHAP are both good approaches to explain models. In the case of fraud detection systems, the main concerns for the selection of an explanation method are the trade offs between model complexity, explanation reliability and run time. In Table~\ref{tab:runtime} we documented that LIME is much faster in providing single instance explanations than SHAP. However, this can come at the cost of reliability. Fraud datasets contain mixed types of variables with many categorical, numerical and text data. In cases where there are a lot of categorical variables, like in the fraud dataset used in this paper, using LIME results in high discrepancies between the predicted value and the explanation because of the perturbation the method used to create the approximate model. We found that one of the combinations that provided the best trade off was to select SHAP with a background dataset of sample size 600. In this case, the mean prediction of the explanation was closest to the mean value of the model while the run time overhead was close to that of LIME.

\section{Challenges and Opportunities}

The development of explainable machine learning methods still faces some research, technical and practical challenges, particularly in anomaly detection methods. One of the main challenges are largely imbalanced datasets. The benign transactions influence explanations more than the fraudulent ones. SHAP can help by specifying the background dataset or reference point depending on the goal of the explanation.

Besides, to be able to trust the explanations received, we need to be able to evaluate them. Evaluation, however, has proven to be a challenging point for explanations. One way to evaluate an explanation is by checking if the explanation was sufficient to allow an end user to achieve their task (i.e. in the fraud example the explanation should include enough details and features that will allow a fraud analyst to effectively decide if the transaction was flagged correctly). It is equally important to be able to identify situations where the AI system is not performing as it should and human intervention or collaboration between human experts and AI is needed. To create effective human computer collaboration, we need to understand and be transparent about the capabilities and limitations of the AI system~\cite{amershi2019guidelines}. 

Another challenging aspect of particular interest to the financial domain is the confidential and private nature of the data. Financial datasets contain sensitive personal and corporate information that should be protected. This usually means that the datasets are anonymized and even the feature names can be changed in the format of `M1' or `id-1', as we have seen in the case study. In these cases it is very difficult to explore the data, a lot of time needs to be spent on masking, unmasking, reverse engineering and deciding whether to include or exclude confidential features since the model cannot be transparent about them and provide explanations that are understandable to an end user.

There are also contextual factors that need to be taken into account when presenting explanations. In a perfect case, the explanations provided by a machine learning method would be identical to human understanding and match with the ground truth. However, that is not usually the case: the explanations provided by the methods analyzed in this paper might be understandable to expert data scientists, but might not be as easy to understand by fraud analysts or end users. How an explanation is presented to the end user (e.g. visualizations, textual explanations, numerical, rule-based or mixed approaches) can determine how effective the explanations are in helping the user understand the inference process and the output of the model. 

A common issue in fraud detection methods is a large false positive rate (i.e. transactions falsely classified as fraud). We can use explanations to verify whether the features in anomaly detection are indeed making sense and are what we would expect. Explaining an anomaly detection model in critical domains is equally important with the model’s prediction accuracy, as it enables end users to understand and trust these predictions and act upon them. One of the main advantages of explaining anomalies is being able to differentiate between detecting fraudulent anomalies and detecting rare but benign events, that could be domain-specific, from genuine users. By presenting explanations for the outliers found in financial fraud detection we can reduce the time and effort needed by fraud analysts to manually inspect each case. 

Furthermore, it is common in fraud detection to experience data shifts (e.g. changing spending patterns in certain times of the year or due to unforeseen circumstances like the Covid-19 pandemic). Most fraud detection algorithms rely on unsupervised learning or anomaly detection and reinforcement learning. In cases like these, we cannot be sure what the algorithm learns because the data shifts can lead to concept drifts (i.e. the model predicts something different than its original purpose). Explainable AI can indicate if there is any data or concept drift of the model. It also makes it easier  to improve and debug the models, and to reuse the models without having to learn and figure out the biases from the start each time they are updated.

Another case where explainable AI could help in fraud detection is adversarial behavior. In adversarial machine learning, an adversary inserts specific instances in a machine learning model knowing that it will affect its learning and cause it to misclassify certain instances (i.e. a cyber criminal could insert perturbed instances in the dataset and affect the fraud score assigned to transactions).
Explainable AI is one means to enhance protection against adversarial attacks. Detecting such attacks is not trivial and explainable AI can have a great impact in assisting in the detection of such manipulation, giving companies and end users more trust in the machine learning inferences.

\section{Conclusion}
In order for advanced machine learning algorithms to be successfully adopted in the financial domain, model explainability is necessary to address regulatory requirements and ensure accurate results. The relevant literature proposed several methods to detect and explain anomalies in different settings from network traffic to insurance fraud. However, we found that exploration of the reliability and practical considerations of real time systems was limited. In this work, we provide insights on trade offs for explanations of financial fraud decisions. We extend current literature by exploring different reference points and comparing the performance of methods for real time fraud systems. Using a transparent Logistic Regression model as ground truth, we find that attribution methods are reliable but can be sensitive to the background dataset, which can lead to different explanation models. Thus, choosing an appropriate background is important and should be based on the goals of the explanation. 

We also found that while SHAP gives more reliable explanations, LIME is faster. In real time systems it is not always feasible to explain everything. We must balance the deployability of the models and explanation methods with the time needed for a human and the likelihood of fraud. It may be beneficial to use a combination of both methods where LIME is utilized to provide real time explanations for fraud prevention and SHAP is used to enable regulatory compliance and examine the model accuracy in retrospective.

\bibliographystyle{splncs04}
\urlstyle{same}
\bibliography{references}

%\printbibliography
%

\end{document}